\documentclass[twoside,11pt]{article}

\usepackage{jmlr2e}

\usepackage[table]{xcolor}
\usepackage[flushleft]{threeparttable}
\usepackage{adjustbox}
\usepackage{cellspace}
\usepackage{amsmath}
\usepackage{caption}

\ShortHeadings{ICML 2020 Workshop on Real World Experiment Design and Active Learning}{Workshop on Real World Experiment Design and Active Learning}
\firstpageno{1}

\begin{document}

\title{Experimental design for bathymetry editing} 

\author{\name Julaiti Alafate \email jalafate@ucsd.edu \\
    \name Yoav Freund \email yfreund@ucsd.edu \\
    \addr Department of Computer Science and Engineering \\
    University of California San Diego \\
    La Jolla, CA 92093, USA
    \AND
    \name David T.\ Sandwell \email dsandwell@ucsd.edu \\
    \name Brook Tozer \email btozer@ucsd.edu \\
    \addr Scripps Institute of Oceanography \\
    University of California San Diego \\
    La Jolla, CA 92093, USA
}

\editor{}

\maketitle

\begin{abstract}
  We describe an application of machine learning to a real-world computer
  assisted labeling task. Our experimental results expose significant
deviations
  from the IID assumption commonly used in machine learning. These
  results suggest that the common random split of all data into
  training and testing can often lead to poor performance.
\end{abstract}

\begin{keywords}
IID assumption, Experimental design, Boosting
\end{keywords}

\section{Introduction}
We present results from a large-scale computer assisted labeling
problem.  We use boosted decision trees as our learning engine,
specifically the LightGBM package~(\cite{ke_lightgbm:_2017}).  We
use the normalized margin as a measure of prediction
confidence~(\cite{schapire1998boosting}).

The standard experimental design for batch learning is to split the
data at random into the train and test sets. {\bf The main contribution of
  this paper} is to show that in our case, this design leads to poor
performance. We show alternative designs that perform better. We argue that
this is not an isolated case and that non-standard experimental design
are likely to perform better in many situations.

\section{Computer assisted bathymetry editing}

\begin{figure}[ht]
  \centering
  \includegraphics[width=1.0\linewidth]{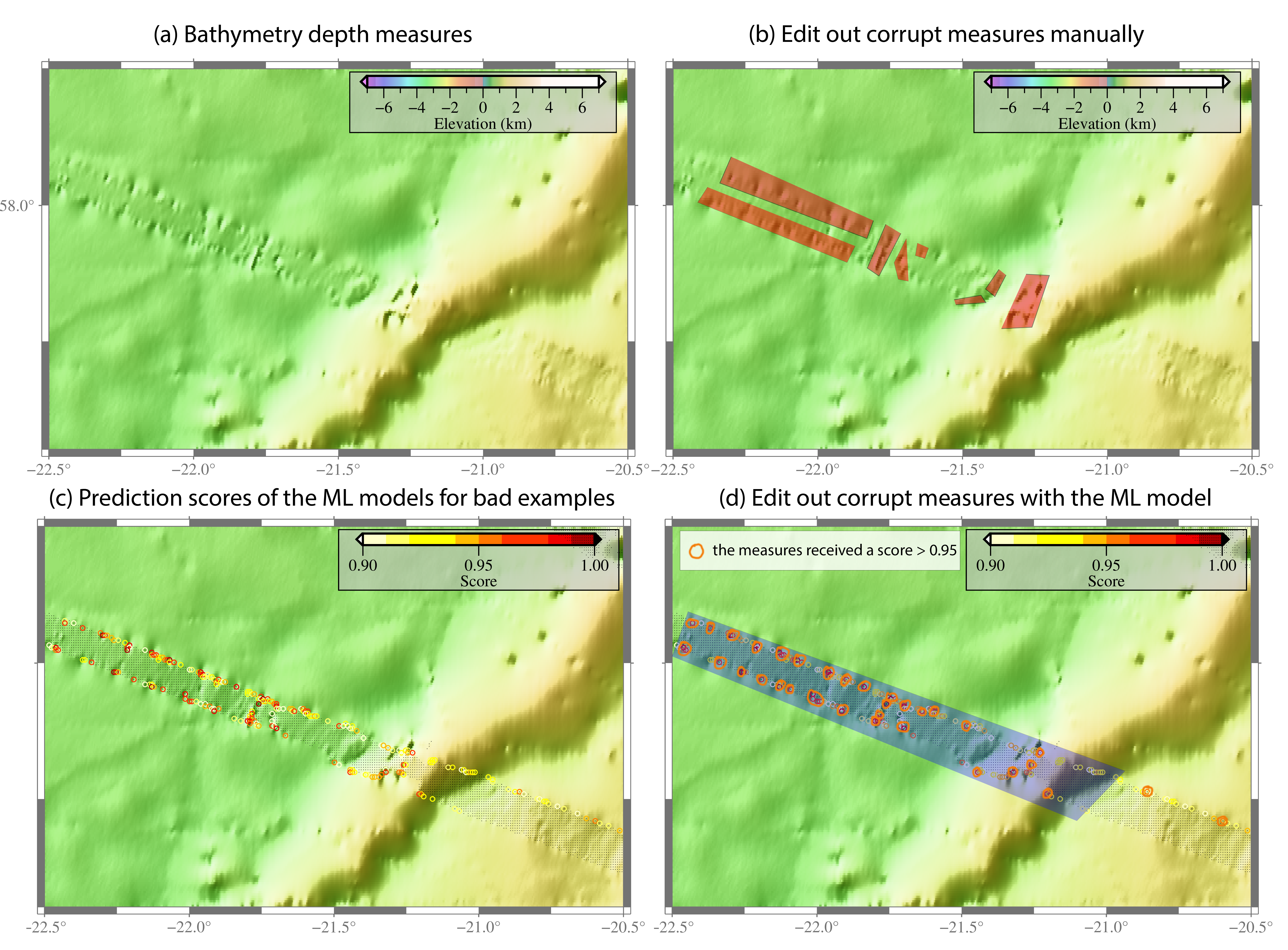}
  \caption{  \label{fig:scores-nopa}
    Computer-assisted bathymetry editing: {\bf (a)} A segment
    of ocean sea floor which includes bad measurements. {\bf (b)}
    Manual editing requires drawing small polygons to identify the
    measurements to be removed. {\bf (c)} Scores generated by the
    boosted trees algorithm. {\bf (d)} Computer-assisted editing, in which the
    editor draws a larger rectangle and specifies a minimal threshold
    for the measurements to be removed.}
\end{figure}

Bathymetry is the mapping of the topography of the oceans
floor~(\cite{tozer2019global}).  Modern bathymetry drawn on two sources
of information: satellite altimeters that have global coverage but
poor resolution, and shipboard echo sounders, which provide
potentially more accurate data only along the path of the ship. The
quality of the data from the echo sounders varies widely. As a
consequence, extensive quality assurance is performed on the data
after it is collected from the ships and before it is integrated into
the map. This process is called ``editing'' and corresponds to assigning
a binary label to each measurement. Measurements labeled ``good'' are
incorporated into the bathymetry map while ``bad'' measurements are
discarded. This process has traditionally been done manually by domain
experts, who
analyze the map after a full update (using all measurements), then
identify and remove suspicious outliers in the updated
map~(Figure~\ref{fig:scores-nopa}\,(a,b)).

Manual editing is labor intensive and the accuracies of the results vary
from person to person. The goal of our project is to develop a
``computer-assisted editing'' method in which the computer assigns a
confidence-rated prediction to each measurement. The human editor can
then select a large area and instruct the computer to remove all bad
measurements within that area~(Figure~\ref{fig:scores-nopa}\,(c,d)).

The learning engine we use is boosted trees, as implemented in
LightGBM. Our confidence score is based on the margin
theory~(\cite{schapire1998boosting}) and is similar to previous work on
boosting-based active learning.

Our initial results were very
good, in fact, unbelievably good. It took some work to identify the
reasons for this overly optimistic results. Those reasons are the
subject of this paper.

\section{Bathymetry data}
\label{sec:bathymetry}

Bathymetry is collected by ships
worldwide~(\cite{tozer2019global}). Each ship collects data on behalf of
one of 17 organizations (the organizations often concentrate on different
geographical regions, thus in this paper we use the term ``organization''
and ``region'' interchangeably).
The combined number of
measurements is over 400 millions and growing.  The raw data size is
203 GB on disk.
The data are not uniformly distributed across the regions. Smaller regions
contain about 4 million measurements, while larger regions contain over 100
million measurements.

The quality of data from different regions varies widely due to type
of ship, reliability of the ship's crew, and complexity of the seafloor
topography of the measured regions.  The fraction of bad measurements
per region varies from less than 0.01\% to 13.12\%.  In addition, the
quality of the editing process varies significantly among the regions
due to the incorrect data labeling by less attentive human editors.

\section{Challenges to the IID assumption}
\label{sec:NotIID}

In general, the goal of batch learning is to generate classifiers that
can accurately classify {\em new} test data that was not
available at training time. For this to be possible the training data
and test data have to be linked in some way. The most common
assumption linking the training and testing data is that the examples
(in both sets) are {\em Identically and Independently Distributed
  (IID)}. This assumption justifies a common experimental design:
collect as much data as possible, randomly permute the data, and
partition it into a training set and a test set. Use the
training set to train the model and use the test set to test
it. Under the IID assumption the test error is an unbiased estimate of
the true generalization error.

We started our project using the random split design, then we found out
several ways that it fails. We describe two failure modes which
contradict the IID assumption. The first, which we call {\em
  sequentiality} contradicts the \emph{Independence} assumption, the second,
which we call {\em diversity} contradicts the \emph{Identical Distribution}
assumption.

\begin{figure}[t]
  \centering
  \includegraphics[width=1.0\linewidth]{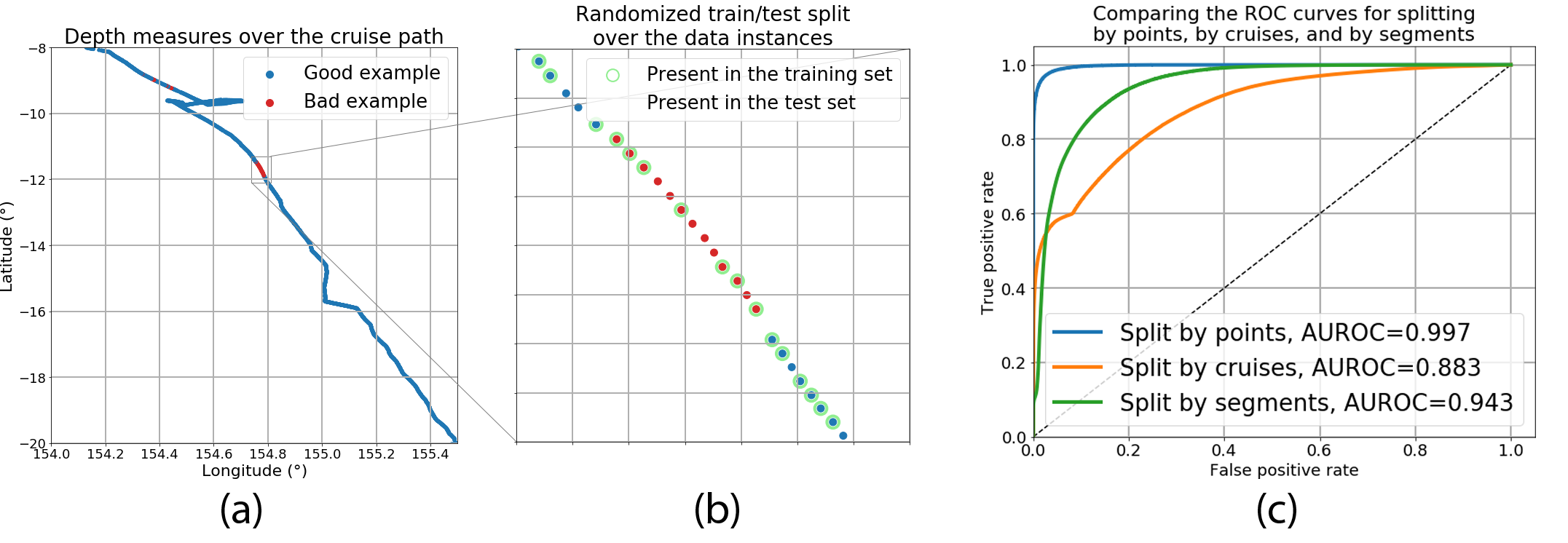}
  \caption{An example of the sequentiality problem in
    bathymetry. {\bf (a)} is the route of a typical ship cruise. {\bf
      (b)} is a zoomed-in part of the route, depicting the partition
    of examples into training and testing. {\bf (c)} contain the ROC
    curves for three different ways of partitioning the data into
    train and test: individual examples, whole cruises, and segments
    of 100,000 measurements.
  \label{fig:sequences}
  }
\end{figure}

\subsection{Data sequentiality} 

Our first set of experiments followed the standard design: the
measurements were partitioned at random between the training set and
the test set. The results on both the training set and the test set
were spectacular, with an area under the test set ROC of
$0.9975$. However, the accuracy fell off significantly when the
classifier was used on a cruise that was not in the training set.

We discovered that this was a result of the way in which the data was
split into a training set and a test set. The standard train/test
partitioning places each {\em individual measurement} in the training
set or the test set respectively. On the other hand, as shown in
Figure~\ref{fig:sequences}, label sequences produced by humans are
likely to have long stretches of good or bad elements. As a result,
each test example is likely to be labeled in the same way as its
neighboring training examples. In other words the examples are
statistically dependent, breaking the independence part of the IID assumption.

Partitioning individual measurements into a train set and a test set
results in a situation where it is sufficient for the classifier
to learn the boundaries of the bad segments and ignore the meaningful
features. This is possible even if the training set is shuffled
because time of day, latitude, or longitude can serve as proxies for
the location of the element in the sequence. The resulting classifier
has a very high test AUROC, but performs poorly on unseen cruises.

To mitigate this problem we changed the way we partition the data into
training and testing. Namely, we divided the data sequences into long
sub-sequences and then place each sub-sequence at either the training
set or the test set at random. We experimented with two ways of
forming sub-sequences. The first way is to take whole cruises as
sub-sequences, the second is to take non-overlapping sub-sequences of
100,000 measurements, which correspond to traveling about 2500 kilometers for the
multi-beam cruises. Most cruises are shorter than 100,000 and are
included as they are, but some are much longer and are split into chunks.
Partitioning the sequences
in these ways reduces the AUROC significantly, as is shown in
Figure~\ref{fig:sequences}. However, this performance is a better
predictor of the future performance.

\begin{figure}[ht]
\begin{minipage}[b]{0.49\textwidth}
    \centering
    \includegraphics[width=1.0\linewidth]{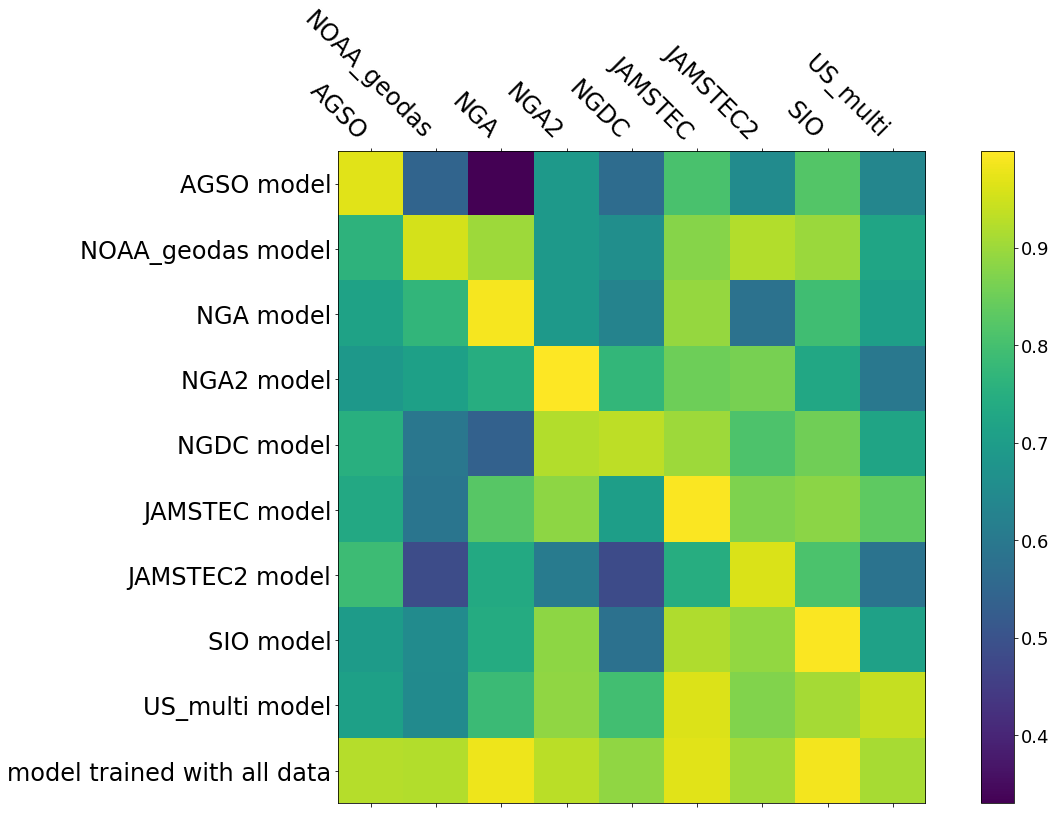}
    \caption{Area under ROC (AUROC) for the models trained using data from one
region and tested on the data from another region. Higher values are better.}
    \label{fig:matrix}
\end{minipage}
\hfill
    \begin{minipage}[b]{0.49\textwidth}
      \centering
  \begin{tabular}{| c | c |}\hline
  \textbf{Region} & \textbf{Improvement} \\ \hline
  AGSO & 4.29\% \\ \hline
      NOAA\_geodas & 3.38\% \\ \hline
      NGA & 0.78\% \\ \hline
      NGA2 & 6.94\% \\ \hline
      NGDC & 4.38\% \\ \hline
      JAMSTEC & 2.59\% \\ \hline
      JAMSTEC2 & 5.57\% \\ \hline
      SIO & 0.73\% \\ \hline
      US\_multi & 2.95\% \\ \hline
  \end{tabular}
  \captionof{table}{Improvement in terms of AUROC of a model trained on
the data sampled from the same region as the test data over the model trained
on all of the data}
      \label{tbl:all2one}
\end{minipage}
\end{figure}

\begin{figure}[ht!]
  \centering
  \includegraphics[width=1.0\linewidth]{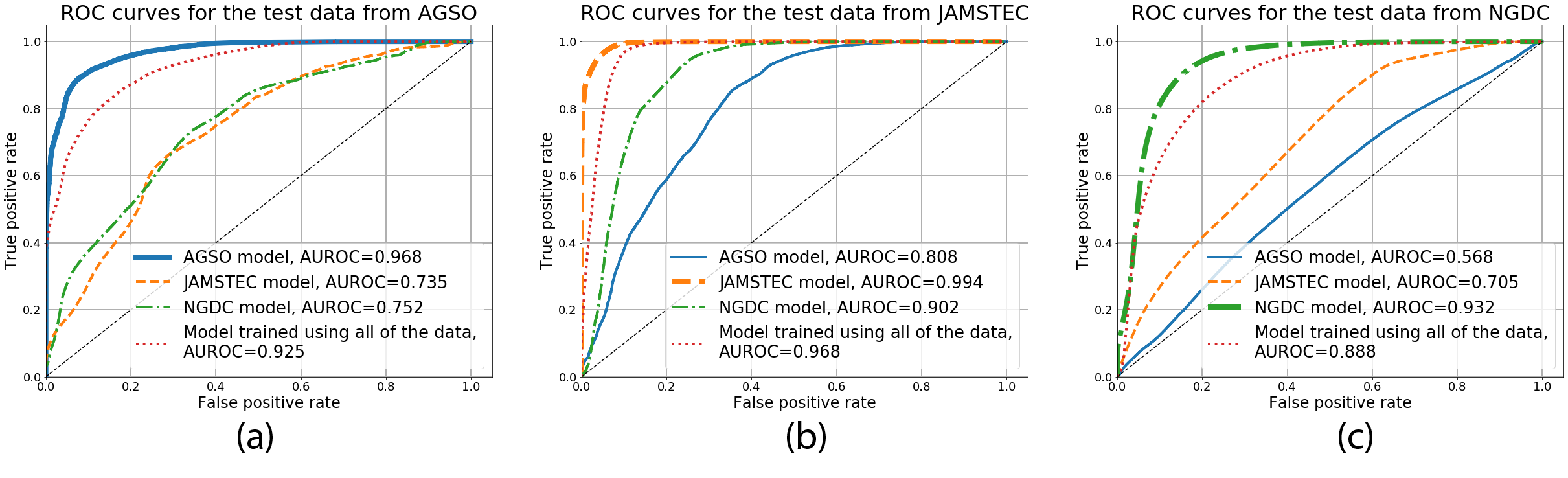}
  \caption{Comparison of the test performance using the models trained using
the data from different sources. The plots show the test performance on the
test data sampled from the
datasets provided by \emph{AGSO} (Australia), \emph{JAMSTEC} (Japan), and
\emph{NGDC}
(US-based data center with global data coverage).
In all three cases, the optimal models are the one trained on the data sampled
from the same origin
as the test data.
The model trained using all of the data is also included for comparison.}
  \label{fig:diversityROC}
\end{figure}

\subsection{Data diversity}

After sequentiality, the main problem we found with the analysis of
bathymetry data is that of data diversity. The distribution of
measurements, and of errors in measurement, depends on many
factors. At the cruise level: The ship and the crew of the ship,
whether bathymetry is the main goal of the cruise, sea conditions
etc. At the level of region, or the organization to which the ship
belongs, there are matters of policy, planning and the resources put
towards quality assurance and data editing.

As a result of this diversity, data collected in different regions has
significantly different distribution. Each row of the matrix in
Figure~\ref{fig:matrix} corresponds to a model (classifier) trained
using data from one of the 17 regions, the bottom row correspond to
the model trained using all of the data. The columns correspond to the
region from which the test set is taken (train and test sets from each
region are randomly partitioned over subsequences).
The colors correspond to correspond
to the AUROC when the row model is tested on the column test
data. First, observe that some pairs, such as model \emph{AGSO} and test
\emph{NGA},
or \emph{JAMSTEC} model and test \emph{NGDC} have a very poor AUROC of 0.5 or
smaller. In other words, the distributions corresponding to different
regions are very different, which is why we say the data source of the bathymetry
dataset is diverse. This contradicts the identically distributed part of the IID
assumption.

Further, the highest AUROC in each column corresponds to the the model
trained on data from the same region. The second best is usually the
classifier trained using all of the data. However, as shown in
Table~\ref{tbl:all2one}, the classifier using just the data
corresponding to the the same region is, in all cases, better than the
classifier using all of the data.

We show the full ROC curves of three specific regions in Figure~\ref{fig:diversityROC}.
We sampled the test sets from \emph{AGSO} (Australia-based organization),
\emph{JAMSTEC} (Japan-based research institution), and \emph{NGDC} (US-based data
center). We then used these test sets to evaluate four models:
first three are trained using the data from each one of the three regions,
and the fourth one trained using all of the data.
In all three cases, the best performance (in terms of AUROC)
is achieved by the model trained using the
data from the same source as the test data, while the performances of the models trained
using the data from difference sources are significantly worse.
Finally, the performance of the model trained using all of the data falls in between.

We conclude that for the bathymetry editing task, combining all data
into one large training set is inferior to using only data that is
similar to the data in the test set. In other words, maximizing the
size of the training set should be balanced against the similarity of
the training set to that of the expected test set.

\section{Summary}

We provide experimental evidence that bathymetry data does not obey
the IID assumption usually made in machine learning. We draw two
conclusions regarding the experimental design of machine learning. 
First - data gathered sequentially should be treated in a way that
ensures that dependencies between neighboring examples do not bias the
generated classifier. Second - when a data source is diverse, it is
not enough to collect large amounts of data. One also needs to ensure
that the diversity of the training set represents the expected
diversity of future test data.

\acks{We would like to acknowledge support for this project
from the National Institutes of Health (NIH grant U19 NS107466)
and the Scripps Institution of Oceanography, UC San Diego. }


\vskip 0.2in
\bibliography{references}

\end{document}